\documentclass[letterpaper]{article} 
\usepackage[submission]{aaai25}  
\usepackage{times}  
\usepackage{helvet}  
\usepackage{courier}  
\usepackage[hyphens]{url}  
\usepackage{graphicx} 
\usepackage{natbib}  
\usepackage{caption} 
\usepackage{algorithm}
\usepackage{algpseudocode}
\usepackage{newfloat}
\usepackage{listings}
\usepackage{amsmath}
\usepackage{xcolor}

\usepackage{multirow}
\usepackage{multicol}
\usepackage{amsfonts}
\usepackage{tcolorbox}
\usepackage{booktabs}
\usepackage{siunitx}
\usepackage{lineno}
\usepackage[export]{adjustbox}

\usepackage{alphalph,etoolbox}

\newcommand{\proj}{LKD4DyTAG }
\newcommand{\projshort}{LKD4DyTAG}

\DeclareCaptionStyle{ruled}{labelfont=normalfont,labelsep=colon,strut=off} 
\floatstyle{ruled}
\newfloat{listing}{tb}{lst}{}
\floatname{listing}{Listing}
\title{LLM-driven Knowledge Distillation for Dynamic Text-Attributed Graphs}
\author {
    Amit Roy\textsuperscript{\rm 1}\thanks{Corresponding Author | Work done as an intern at Futurewei Technologies Inc.} ,
    Ning Yan\textsuperscript{\rm 2},
    Masood Mortazavi\textsuperscript{\rm 2}
}
\affiliations {
    \textsuperscript{\rm 1}Purdue University,
    \textsuperscript{\rm 2}Futurewei Technologies  Inc.\\
    roy206@purdue.edu, 
    \{yan.ningyan, masood.mortazavi\}@futurewei.com
}

\usepackage{bibentry}

\begin{document}

\maketitle

\begin{abstract}
\textbf{Dy}namic \textbf{T}ext-\textbf{A}ttributed \textbf{G}raphs (\textbf{DyTAG}s) have numerous real-world applications, e.g., social, collaboration, citation, communication, and review networks. In these networks, nodes and edges often contain text descriptions, and the network structure can evolve over time. 
Future link prediction, edge classification, relation generation, and other downstream tasks on DyTAGs require powerful representations that encode structural, temporal, and textual information. Graph Neural Networks (GNNs) are adept at managing structural data, yet encoding temporal information in dynamic graphs has proven challenging. 
In this work, we propose \textbf{L}LM-driven \textbf{K}nowledge \textbf{D}istillation for \textbf{Dy}namic \textbf{T}ext \textbf{A}ttributed \textbf{G}raph (\textbf{\projshort}) with temporal encoding to address those challenges. 
We first use a simple yet effective approach to encode temporal information in edges so that graph convolution can simultaneously capture both temporal and structural information in the hidden representations. 
%
%
To leverage LLM's text processing capabilities to learn richer representations on DyTAGs, we distill knowledge from LLM-driven edge representations (based on a neighborhood's text attributes) into saptio-temporal representations learned by a lightweight GNN model that encodes temporal and structural information. Our knowledge distillation objective enables the GNN to learn representations that more effectively encode available \textit{structural}, \textit{temporal} and \textit{textual} information in DyTAGs.
Extensive experimentation conducted on six real-world dynamic text-attributed graph datasets prove the efficacy of our approach \proj for the future link prediction and edge classification task. The results show that our approach significantly improves the performance of downstream tasks compared to baseline models.

\end{abstract}

\begin{figure}
\centering
\includegraphics[width=0.5\textwidth,left]{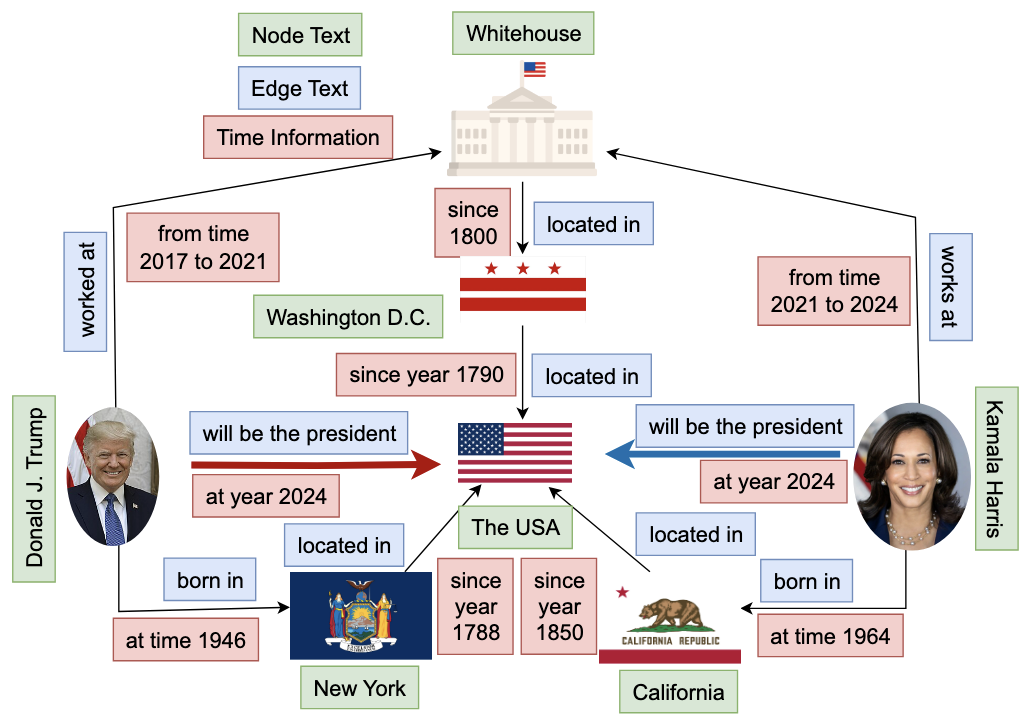}
    \caption{\fontsize{9}{10} \selectfont Example Dynamic Text-Attributed Graph in the context of 2024 Presidential Election in USA. Each entity is described with text, while the relations between them are represented by edges including temporal information.}    \label{fig:DyTAG_example_election_2024}
    \vspace*{-1em}
\end{figure}

\section{Introduction}

\textbf{Dy}namic \textbf{T}ext-\textbf{A}ttributed \textbf{G}raph (\textbf{DyTAG}) structures change over time, incorporating text descriptions in both nodes and edges. DyTAGs have a wide range of applications across various domains~\cite{tang2023dynamic,luo2023hope,zhang2023streame, cai2022temporal,skarding2021foundations, kazemi2020representation}, such as email networks, political event networks, online Q\&A forums, and review networks for products and movies. 
For instance, in an email network, nodes can represent email users with their profile information, while edges can represent the emails exchanged between them at specific times. 
Similarly, in a movie or item review network, nodes can describe users, movies, or items, while edges may denote reviews between users and movies or items at different times. 
Again, in a political event network nodes can denote the political figures, places, institution and the edges may indicate the interrelationships between those entities with temporal information attached. 
Handling both the dynamically changing graph structure and the semantic information in nodes and edges described in natural language is a challenging task since structure and text have different modality. When designing deep learning models for various downstream tasks for such text attributed dynamic graph data, it is crucial to carefully consider structural, temporal, and textual information.

Dynamic networks are often represented as snapshots with discrete timestamps.
Recent works propose various methods to capture temporal information in dynamic networks, such as anonymous random walk-based approach~\cite{wang2021inductive} and window-based approach~\cite{alomranidyg2vec}. 
However, when the number of timestamps is extensively large and the downstream task depends on earlier timestamps, it becomes challenging to efficiently encode the complete temporal information into the latent representation. Additionally, as the graph structure evolves over time, it is also essential to include temporal information in graph convolution as well as structural information. 
In addition to temporal and structural information found in dynamic graphs, Dynamic Text-Attributed Graphs(DyTAGs) also incorporate textual information. 
%
%
This added text provides a more comprehensive representation of real-world systems compared to dynamic graphs that only contain structural and temporal data.

With the increasing capability of LLMs to produce semantic representations of text, numerous works have emerged that utilize them for graph tasks~\cite{fang2024gaugllm, xu2024llm}. 
Most of these works rely on prompt engineering to describe the graph structure and the downstream task in order to ``tune'' the LLMs, expecting the LLMs to generate the desired results for the given task.
However, despite their advanced capabilities, LLMs face challenges in scalability and are limited in their ability to process input prompts of longer sequence lengths. Consequently, their applicability to diverse downstream tasks is constrained.  
Furthermore, despite LLMs versatility in capturing semantic information from text, they often try to memorize the common structural motifs/patterns and are not effective in comprehending the underlying evolving structure and the complex temporal dynamics of Dynamic Text-Attributed Graphs (DyTAGs). 
%
%
To address these challenges, one can deploy the capabilities of LLMs for scalable processing of all locally-scoped text attributes in DyTAGs and transfer representations thus learned by heavyweight large language models to lightweight graph neural networks models (GNNs) for downstream tasks involving DyTAGs. 
Here, text-based edge representation generated by the LLM serves as a guide for the spatio-temporal edge representation produced by temporally encoded GNNs. This approach ensures that, among the three types of information present in DyTAGs, temporal and structural information are captured by the temporally encoded GNN, while the LLMs are utilized to encode the textual information of edges for distillation purposes.

This approach leads to our design of a novel LLM-driven knowledge distillation framework for processing
Dynamic Text-Attributed Graphs (DyTAGs) with temporal encoding.
The structure and temporal information of edges are transformed into spatio-temporal edge representations using GNNs with a simple yet efficient encoding of time. 
For the textual edge representation, we describe the neighbors of adjacent nodes to the LLMs one edge at a time and sum these descriptions to obtain neighborhood textual embeddings. 
Next, we add up the neighborhood textual embeddings of adjacent node with the LLM output for the corresponding edge's description to obtain a text-based edge representation.
Next, we bring into alignment the spatio-temporal edge representation and the text-based edge representation. 
In short, we are using knowledge distillation method to transfer knowledge from the text-based edge representation from a teacher LLM model to the spatio-temporal representation of a student GNN model with temporal encoding for dynamic text attributed graph. 
Our contributions in this work are as follows:

\begin{itemize}
    \item We propose a novel framework \textbf{\proj} which integrates time, structure, and text information in Dynamic Text-Attributed Graphs to produce suitable representations for downstream tasks.
    \item We designed a simple yet effective approach to produce spatio-temporal edge representations of DyTAG that capture the temporal and structural information using temporal edge encoding with GNNs.
    \item To make the best utilization of textual information from DyTAG and the text processing capability from LLMs, we perform knowledge distillation from text-based edge representation to spatio-temporal edge representation.
    \item We perform extensive experiments on six real-world benchmark dynamic text-attributed graph datasets and show that our proposed method \proj can outperform state-of-the-art approaches in certain downstream tasks such as \textit{future link prediction} and \textit{edge classification}.
\end{itemize}




\begin{figure*}
    \centering
    \includegraphics[width=1\linewidth]{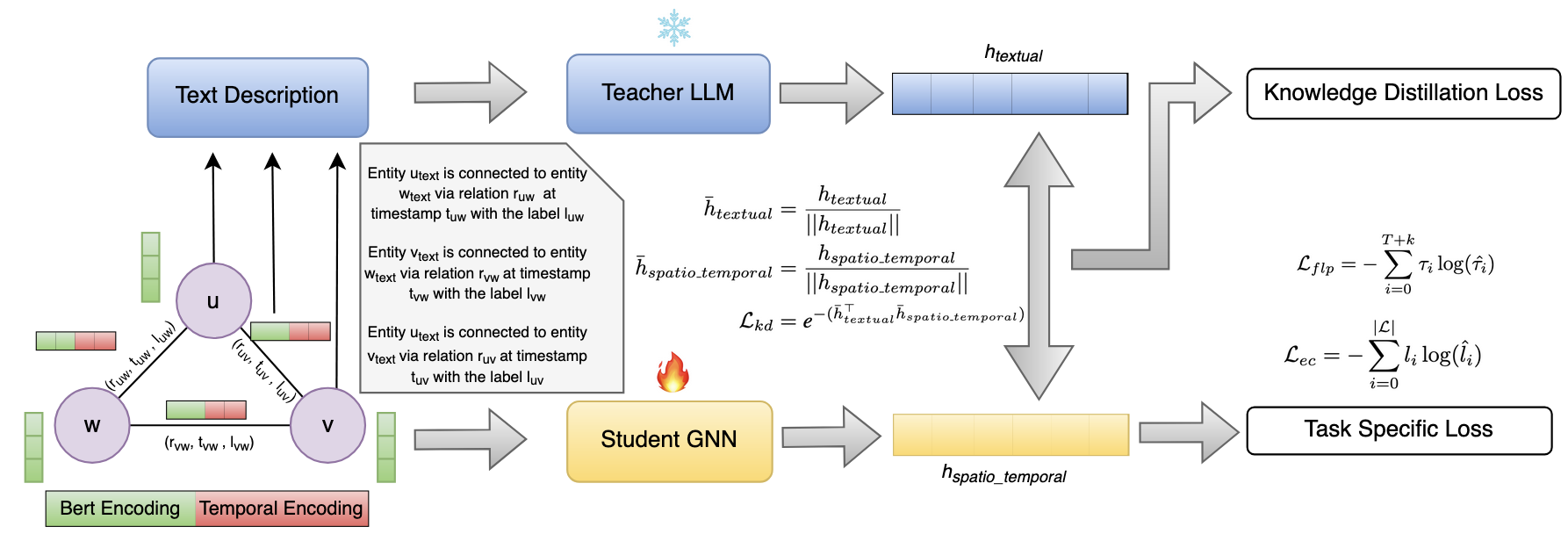}
    \caption{\fontsize{9}{9} \selectfont \textbf{\proj: LLM-driven Knowledge Distillation for Dynamic Text-Attributed Graph with temporal encoding.} \\ \footnotesize{First, temporal encodings (red) are added to the graph edges along with their BERT encodings (green). The graph information is then transformed into textual information by incorporating the adjacent node's 1-hop neighbors to encode the semantic context, as illustrated in the textbox. The derived text description is fed into a pretrained teacher LLM model to obtain the textual representation, while the graph is fed into a trainable student GNN to obtain the spatio-temporal representation. These two representations are brought closer in latent space by minimizing the cosine similarity, which defines the knowledge distillation loss. Additionally, the student model's spatio-temporal edge representation is utilized for the downstream task, which is trained using the task-specific loss.
}}
    \vspace*{-1em}
\label{fig:LLMKD4DyTAG_framework}
\end{figure*}

\section{Notation and Problem Formulation}

\textbf{Dynamic Text-Attributed Graph (DyTAG).}
A Dynamic Text-Attributed Graph (DyTAG) can be defined as $\mathcal{G} = \{\mathcal{V},\mathcal{E}\}$, where $\mathcal{V}$ denotes the set of nodes or entities and $\mathcal{E}$ denotes the set of edges or relations among those entities, each with associated timestamp, relation, and label information. 
The terms node-edge and entity-relation are used in regular graphs and knowledge graphs respectively. We will use both terminologies interchangeably.
Each entity $u \in \mathcal{V}$ has an associated text description $u_{\text{text}}$ that describes the entity. 
Each edge $e_{uv} \in \mathcal{E}$ between two adjacent entities $u$ and $v$ can be described as a three-tuple $\{r_{uv}, t_{uv},l_{uv}\}$, where $r_{uv}$ describes the relationship between $u$ and $v$ in texts as $r_{text}$, $t_{uv}$ describes the timestamp information when $u$ and $v$ are connected, and $l_{uv}$ is the label of the edge $e_{uv}$. 
Each timestamp $t_{uv} \in \mathcal{T}$, and labels $l_{uv} \in \mathcal{L}$, where $\mathcal{T}$ and $\mathcal{L}$ denote the set of timestamps and edge labels, respectively. 
We denote the DyTAG until timestamp $T$ as $\mathcal{G}_{T} = (\mathcal{V}_T,\mathcal{E}_T)$. Additionally, we describe the $1$-hop neighborhood of a node $u$ as $\mathcal{N}_u$. 
In this work, we aim to design a framework that effectively utilize the temporal, structural, and textual information of DyTAGs to address two tasks, \textit{future link prediction} and \textit{edge classification}. 

\textbf{Future Link Prediction.}  Given a DyTAG $\mathcal{G}_{T}$ containing the interactions between nodes and edges with their text descriptions until timestamp $T$, the task of future link prediction aims at predicting the existence of an edge $e_{uv}$ between node $u$ and node $v$ at the next $k$ timestamps $T+k$. For example, given a email network among the users, the task of future link prediction aims to determine whether two users will exchange emails in the near future based on their past email interactions and the contents of the emails.

\textbf{Edge Classification.} Given a DyTAG $\mathcal{G}_{T}$ with the interactions between entities and relations up to timestamp $T$, the task of edge classification focuses on determining the category of an edge between two candidate entities $u$ and $v$ at the future timestamp $T+k$. For example, in an item-review network containing user and item nodes with edges describing reviews between them, the edge classification task involve determining the type of review a user has given for an item, such as a product or a movie.

\section{Methodology}

\textbf{Motivations.} A Dynamic Text-Attributed Graph encompasses structural, temporal, and textual information to describe the dynamic inter-relationships between entities. 
We design \proj utilizing knowledge distillation approach to encode the semantic understanding of a teacher LLM regarding edge creation and edge types from the text description of network elements of an DyTAG into the representation of a lightweight student GNN model, which captures both structural and temporal information. First,  
we learn two different representations for each edge, one from the student GNN which includes structural and temporal information and the other from LLM that covers textual information. Then, we distill the semantic understanding of the LLM representation into lightweight GNN representation by adopting knowledge distillation mechanism.
We illustrate the model architecture of \proj in Figure~\ref{fig:LLMKD4DyTAG_framework} and provide the pseudocode in Algorithm~\ref{alg:LLMKD4DyTAG} in the Appendix.

\textbf{Spatio-temporal Edge Embedding (Student Model).}
To simultaneously encode structural and temporal information into edge representations, we propose a simple yet effective temporal edge encoding method.
For each edge $e_{uv}$, we use a vector $\tau \in \mathbb{R}^{T+k}$ to represent the status of an edge---its non-existence up to timestamp $t_{uv}$ and its existence thereafter---by setting $\tau[0:t_{uv}]=0, \tau[t_{uv}:T]=1$. 
Intuitively, we incorporate temporal information into the edges by using 0 and 1 that denote the non-existence and existence of an edge, respectively. Furthermore, we mask the future timestamps,  $T$ to $T+k$ during the training time, i.e., $\tau[T:T+k]=-1$.
With the temporal information encoded in the edges, we can apply message-passing Graph Neural Networks (GNNs) to learn edge representations that capture both structural and temporal information.
We denote such embedding as $h_{spatio\_temporal}$. 

First, we obtain the node and edge representations using a multi-layer perceptron over the BERT embeddings of the node and edge text. 

\begin{align}
    h_{u} &= \phi_{mlp\_student}(BERT(u_{text}))\\
    h_{v} &= \phi_{mlp\_student}((BERT(v_{text})))\\
    h_{uv} &= \phi_{mlp\_student}(BERT(r_{text}))
\end{align}

Here, $\phi_{mlp\_student}:\mathbb{R}^{d_{BERT}} \to \mathbb{R}^{d_{student}}$ is a multi-layer perceptron that transforms the output embeddings of node text description $u_{text}$, $v_{text}$, and edge text description $r_{text}$ from a frozen BERT encoder into node representations $h_u \in \mathbb{R}^{d_{student}}$ and $h_v \in \mathbb{R}^{d_{student}}$, as well as edge representation $h_{uv} \in \mathbb{R}^{d_{student}}$, which represents the relation $r_{uv}$ between node $u$ and $v$ at timestamp $t_{uv}$. 
Here $d_{bert}$ and $d_{student}$ denote the dimensionalities of the BERT embeddings and student model embeddings, respectively.

Next, we concatenate $h_{uv}$ with the temporal encoding $\tau$ to incorporate temporal information into edges, resulting in $\hat{h}_{uv}$.
We use a $1$-layer message passing GNN $\psi$, which considers edge attributes to capture structural and temporal information through neighborhood aggregation, producing the node representations $\Bar{h}_u$ and $\Bar{h}_v$. Here, $AGG(\cdot)$ can be any aggregation function that combines the neighborhood node and edge embeddings, while $UPDATE(\cdot)$ function integrates the aggregated neighborhood embedding with the target node embedding. After that, we perform element-wise multiplication (Hadamard product) between adjacent node representations $\Bar{h}_u$ and $\Bar{h}_v$  after the graph convolution to obtain $h_{spatio\_temporal}$ for each edge.

\begin{align}
\Hat{h}_{uv} &= h_{uv} || \tau\\ 
\Bar{h}_{u} &= \text{UPDATE}(h_u, \text{AGG}(\{(h_w,\Hat{h}_{uw}): w \in \mathcal{N}_u\}))\\
\Bar{h}_{v} &= \text{UPDATE}(h_v, \text{AGG}(\{(h_w,\Hat{h}_{vw}): w \in \mathcal{N}_v\}))\\
&h_{spatio\_temporal} = \Bar{h}_{u} \odot \Bar{h}_{v} 
\end{align}

With the edge representation $h_{spatio\_temporal}$ learnt from the temporal encoded graph structure by the message passing GNN, we predict the temporal edge encoding $\Hat{\tau}$ as follows:

\begin{equation}
    \Hat{\tau} = \sigma(\phi_{flp}(h_{spatio\_temporal}))
\end{equation}

where $\phi_{flp}$ is a multi-layer perceptron which maps the spatio-temporal representation into a vector of size equal to the total number of timestamps of interest, i.e., $\phi_{flp} : \mathbb{R}^{d_{student}} \to \mathbb{R}^{T+k}$ and $\sigma$ can be an activation function, such as sigmoid.

We learn the GNN parameters by minimizing the binary cross entropy (BCE) loss $\mathcal{L}_{flp}$ that denote the difference between the original temporal edge encoding and predicted one for the future link prediction task.

\begin{equation}
    \mathcal{L}_{flp} = - \sum_{i=0}^{T+k} \tau_i \log( \Hat{\tau_i}) 
\end{equation}

For the edge classification task, we utilize a multi-layer perceptron $\phi_{ec}$ to transform the $h_{spatio\_temporal}$ into the predicted class probabilities of the edges ,i.e., $\phi_{ec} : \mathbb{R}^{d_{student}} \to \mathbb{R}^{|\mathcal{L}|}$.

\begin{equation}
    \hat{l}_{uv} = \sigma(\phi_{ec}(h_{spatio\_temporal}))
\end{equation}

We minimize the BCE loss between the one hot encoding of the true labels of the edge class label $l$ and the predicted class probabilities $\hat{l}$.

\begin{equation}
    \mathcal{L}_{ec} = - \sum_{i=0}^{|\mathcal{L}|} l_i \log( \hat{l}_i) 
\end{equation}

\begin{figure}[!t]
\includegraphics[width=0.45\textwidth]{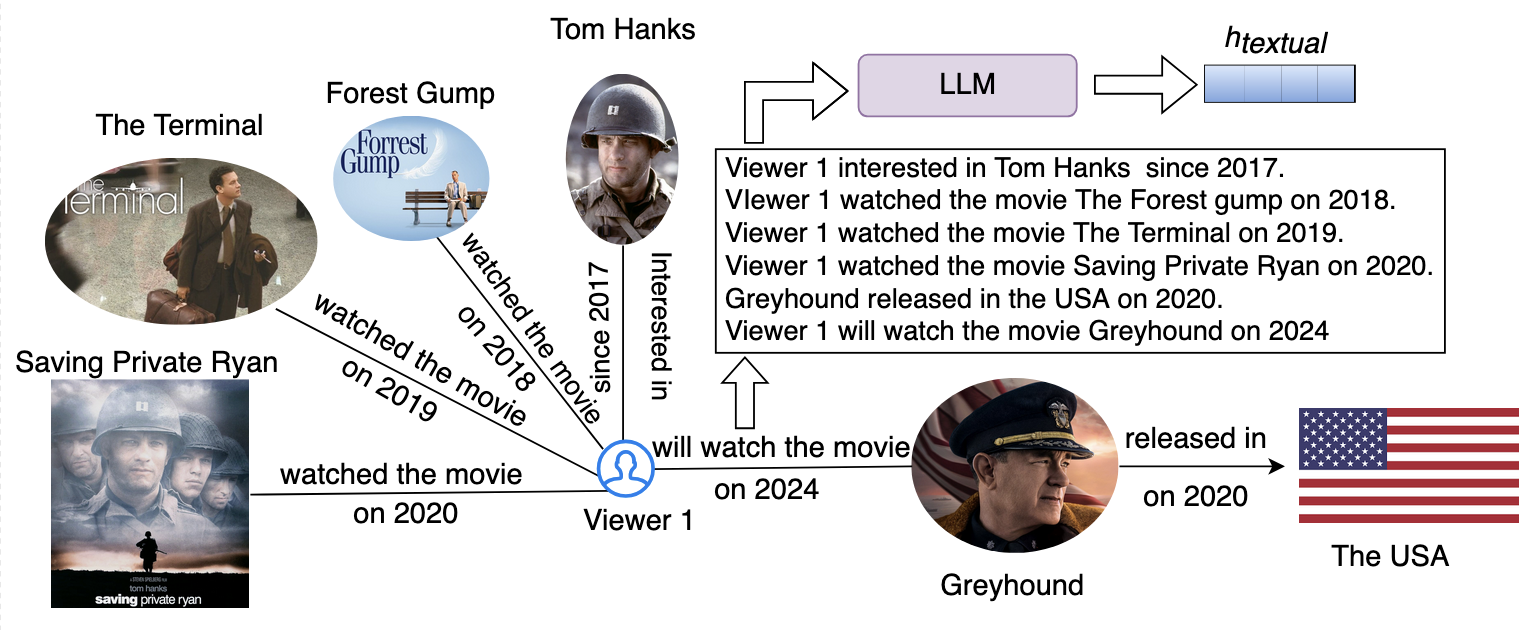}
\caption{\fontsize{9}{9} \selectfont \textbf{Textual edge representation from the LLM using the descriptions of adjacent nodes' neighbors.}}
\vspace*{-2em}
\label{fig:textual_edge_embedding}
\end{figure}

\textbf{Textual Edge Embedding (Teacher Model).} 
Compared to traditional dynamic graphs, DyTAGs are rich in textual information associated with their nodes and edges, necessary for completely describing real-world dynamically changing systems. Therefore, in addition to structural and temporal information, the semantics from the textual information from edge relations, adjacent nodes, and their neighbors is valuable for predicting future links or classifying the label of an edge.

Recently, large language models (LLMs)~\cite{brown2020language} have shown impressive capabilities in understanding text semantics. However, their high storage and computational requirements challenge their direct use in real-time systems like DyTAGs. 
Assuming an LLM can effectively encode edge context into a text-based representation, we aim to use the LLM as a teacher model to distill this knowledge into a spatio-temporal representation by aligning the two in the latent space.
This approach leverages the LLM’s text processing strengths to efficiently encode semantic understanding into dynamic text-attributed graph representations, in a computationally and resource-efficient manner.
Modern LLMs, such as Llama2~\cite{touvron2023llama}, ChatGPT~\cite{brown2020language}, Gemini AI~\cite{reid2024gemini}, excel at generating embeddings that capture text semantics -- crucial for predicting future edge existence and types in link prediction.
We incorporate these capabilities by distilling the LLM's textual edge representation into our spatio-temporal representation. 
We initially obtain the representation of an edge by the describing text of the 1-hop neighborhood of the adjacent nodes and also the edge itself, as illustrated in Figure~\ref{fig:textual_edge_embedding}.

For each true edge or fake sampled edge $e_{uv}$ between two adjacent entities $u$ and $v$ in a particular dynamic text-attributed graph, we design a text prompt by describing the $1$-hop neighborhood of the adjacent nodes $u$ and $v$. We add a decision sentence to denote whether the edge is true or false for future link prediction between entity $u$ and $v$ at timestamp $t$ via relation $r$. Label information for the link prediction task is also included. Note that this latent distilled knowledge objective is not available at test time.
%
%
%



\textbf{LLM-driven Knowledge Distillation.} The LLM's representation can grasp the semantics and context of edge creation and the adjacent nodes. However, obtaining such representation from an LLM using text prompt is resource-intensive and computationally demanding. Therefore, we employ knowledge distillation to transfer the LLM's capabilities to our spatio-temporal representation, which can be obtained in a resource-efficient and scalable manner.

We distill the knowledge from LLM-based representation into spatio-temporal encoding from GNN-based representation by making $h_{spatio\_temporal}$ to be as close as $h_{textual}$ in the latent space.



First, we compute the neighborhood textual embedding $h_{\mathcal{N}(u)} \in \mathbb{R}^{d_{teacher}}$ for each node $u$ by describing each neighboring edge $e_{uw}$ in the node's $1$-hop neighborhood to the LLM, where ${d_{teacher}}$ is the dimensionality of the MLP-transformed textual edge representation.

\begin{align}
 \Bar{h}_{\mathcal{N}(u)} &= \sum_{w \in \mathcal{N}(u)} LLM(neighbor\_prompt(e_{uw})) \\
 h_{\mathcal{N}(u)} &= \phi_{mlp\_teacher}(\Bar{h}_{\mathcal{N}(u)})
\end{align}

Here, $\phi_{mlp\_teacher}$ is an MLP for the teacher model which transforms the combined frozen LLM embeddings into textual representation for each edge, i.e., $\phi_{mlp\_teacher} : \mathbb{R}^{d_{LLM}} \to \mathbb{R}^{d_{teacher}}$ where $\mathbb{R}^{d_{LLM}}$ is the dimensionality of LLM output embeddings. 

Next, for each edge $e_{uv}$ we sum up the neighborhood textual embedding $h_{\mathcal{N}(u)}$ and $h_{\mathcal{N}(v)}$ of the adjacent nodes $u$ and $v$ with the $\Tilde{h}_{uv}$ , the textual link representation for edge $e_{uv}$. 
We use $an/no$ term in the edge description to differentiate between true and fake edges for the link prediction task during the training.
This serves as a mechanism for the LLM to contrast positive and negative edges in the hidden representation.

\begin{align}
 \Tilde{h}_{uv} &= \phi_{mlp\_teacher}(LLM(link\_prompt(e_{uv}))) \\
 h_{textual} &=  h_{\mathcal{N}(u)} +  h_{\mathcal{N}(v)} + 
\Tilde{h}_{uv}
\end{align}

Our assumption is that LLM can produce an appropriate representation of each edge in a DyTAG by comprehending the semantics of edge creation and edge labels since the node and edge texts describe actual relationships between adjacent entities in DyTAGs. 
We aim to align the temporal edge encoding-based spatio-temporal representation from the GNN with the textual edge encoding from the LLM.
This is achieved by minimizing the negative exponential of the similarity between the two embeddings,  thereby bringing $h_{textual}$ and $h_{spatio\_temporal}$ closer by reducing the angle between these representations in the latent space.

We normalize both embeddings $h_{textual}$ and $h_{spatio\_temporal}$. Then, we define our knowledge distillation loss as the negative exponential of the dot product between $\bar{h}_{textual}$ and $\bar{h}_{spatio\_temporal}$, intuitively the similar $\bar{h}_{textual}$ and $\bar{h}_{spatio\_temporal}$ are in the latent space the smaller $\mathcal{L}_{kd}$ is. Note that the LLM is frozen and knowledge distillation occurs through the latent space. 

\begin{align}
    \bar{h}_{textual} &= \frac{h_{textual}}{||h_{textual}||} \\   \bar{h}_{spatio\_temporal} &= \frac{h_{spatio\_temporal}}{||h_{spatio\_temporal}||} \\
    \mathcal{L}_{kd} &=  \textit{e}^{-(\bar{h}_{textual}^\top \bar{h}_{spatio\_temporal})}
\end{align}



Here, $\lambda_{flp}, \lambda_{ec},  \lambda_{kd}$ are the hyperparameters that denote the importance of the future link prediction, edge classification, and knowledge distillation loss. 



\begin{table*}[!ht]
\centering
\scalebox{0.9}{
\begin{tabular}{c|cccccc}
\toprule
\textbf{Dataset} & \textbf{Nodes} & \textbf{Edges} & \textbf{Edge Categories} & \textbf{Timestamps} & \textbf{Domain} & \textbf{Text Attributes} \\ \hline
Enron                                  & 42,711                              & 797,907                             & 10                                            & 1,006                                    & E-mail                               & Node \& Edge                                                                          \\
GDELT                                  & 6,786                               & 1,339,245                           & 237                                           & 2,591                                    & Knowledge graph                      & Node \& Edge                                                                         \\
ICEWS1819                              & 31,796                              & 1,100,071                           & 266                                           & 730                                      & Knowledge graph                      & Node \& Edge                                                                         \\
Googlemap CT                           & 111,168                             & 1,380,623                           & 5                                             & 55,521                                   & E-commerce                           & Node \& Edge                                   \\
Stack elec                             & 397,702                             & 1,262,225                           & 2                                             & 5,224                                    & Multi-round dialogue                 & Node \& Edge                                                                     \\
Stack ubuntu                           & 674,248                             & 1,497,006                           & 2                                             & 4,972                                    & Multi-round dialogue                 & Node \& Edge                                                                                                       \\ \bottomrule
\end{tabular}}
\caption{\centering \selectfont Dataset statistics of Dynamic Text-Attributed Graph from DTGB benchmark~\cite{zhang2024dtgb}.}
\vspace*{-1em}
\label{tab:DyTAG_datasets}
\end{table*}

\begin{table*}[t]
\centering
\scalebox{0.6}{
\begin{tabular}{c|c|c|c|c|c|c|c|c|c|c}
\toprule \textbf{Metric} &\textbf{Task Type} & \textbf{Datasets} & \textbf{JODIE} & \textbf{DyRep} & \textbf{TGAT} & \textbf{CAWN} & \textbf{TCL} & \textbf{GraphMixer} & \textbf{DyGFormer} & \textbf{\proj} \\
\bottomrule 
\multirow{8}{*}{\textbf{ROC-AUC}} &\multirow{4}{*}{\textit{tr}.} &  Enron & 0.9731 ± 0.0052 & 0.9274 ± 0.0026 & 0.9681 ± 0.0026 & 0.9740 ± 0.0007 & 0.9618 ± 0.0025 & 0.9567 ± 0.0013 & \underline{0.9779 ± 0 .0014} & \textbf{0.9887 ± 0.0011}  \\
& & ICEWS1819 & 0.9741± 0.0113 &  0.9632 ± 0.0027 & 0.9904 ± 0.0039 & 0.9857 ± 0.0018 & \underline{0.9923 ± 0.0012} & 0.9863 ± 0.0024 & 0.9888 ± 0.0015 & \textbf{0.9950 ± 0.0014} \\
 & & Googlemap CT & \textit{OOM} & \textit{OOM} & \underline{0.9049 ± 0.0071} & 0.8687 ± 0.0063 & 0.8348 ± 0.0094 & 0.8095 ± 0.0014 & 0.8207 ± 0.0018 & \textbf{0.9127 ± 0.0037} \\
& & GDELT & 0.9533 ± 0.0020 & 0.9453 ± 0.0018 & 0.9595 ± 0.0033 & 0.9600 ± 0.0061 & 0.9619 ± 0.0008 & 0.9552 ± 0.0018 & \underline{0.9662 ± 0.0003} & \textbf{0.9998 ± 0.0001} \\ \cline{2-11}
& \multirow{4}{*}{\textit{in}.}  & Enron & 0.8732 ± 0.0037 & 0.7901 ± 0.0047 & 0.8650 ± 0.0032 & 0.9091 ± 0.0014 & 0.8512 ± 0.0062 & 0.8347 ± 0.0039 & \underline{0.9316 ± 0.0015 } & \textbf{0.9367 ± 0.0011}\\
& & ICEWS1819 & 0.9285 ± 0.0065 & 0.9030 ± 0.0097 & \underline{0.9706 ± 0.0054} & 0.9774 ± 0.0039 & \textbf{ 0.9778 ± 0.0012 } & 0.9605 ± 0.0025 & 0.9630 ± 0.0027 & 0.9543 ±. 0.0013 \\
 & & Googlemap CT & \textit{OOM} & \textit{OOM} & \underline{0.8791 ± 0.0028} & 0.7058 ± 0.0047 & 0.7895 ± 0.0046 & 0.7895 ± 0.0046 & 0.7648 ± 0.0052 & \textbf{0.8857 ± 0.2327}\\
 & & GDELT & 0.8921 ± 0.0065 & 0.8917 ± 0.0007 & 0.9012 ± 0.0011 & 0.8899 ± 0.0082 & 0.9099 ± 0.0022 & 0.8942 ± 0.0035 & \underline{ 0.9206 ± 0.0003} & \textbf{0.9577 ± 0.0012}\\ \hline
 \multirow{8}{*}{\textbf{AP}} & \multirow{4}{*}{\textit{tr}.} &  Enron & 0.9553 ± 0.0051 & 0.9066 ± 0.0076 & 0.9668 ± 0.0026 & 0.9756 ± 0.0008 &0.9603 ± 0.0018 & 0.9559 ± 0.0027 & \underline{0.9804 ± 0.0015} & \textbf{0.9943 ± 0.0001}\\
& & ICEWS1819 & 0.9752 ± 0.0037& 0.9676 ± 0.0026& 0.9908 ± 0.0032& 0.9886 ± 0.0025& \underline{0.9927 ± 0.0012}& 0.9871 ± 0.0034& 0.9901 ± 0.0018& \textbf{0.9970 ± 0.0012}\\
& & Googlemap CT & \textit{OOM} & \textit{OOM} & \underline{0.9002 ± 0.0019} & 0.8721 ± 0.0027& 0.8335 ± 0.0018 &0.8072 ± 0.0010& 0.8183 ± 0.0038 & \textbf{0.9144 ± 0.0013}\\
& & GDELT & 0.9466 ± 0.0032 & 0.9416 ± 0.0017& 0.9572 ± 0.0029& 0.9582 ± 0.0053& 0.9601 ± 0.0011& 0.9523 ± 0.0020& \underline{0.9653 ± 0.0003} & \textbf{0.9776 ± 0.0034}\\ \cline{2-11} 
& \multirow{4}{*}{\textit{in}.} & Enron & 0.8761 ± 0.0023 & 0.7734 ± 0.0044 & 0.8589 ± 0.0031 & 0.9223 ± 0.0011 & 0.8560 ± 0.0024 & 0.8328 ± 0.0034 & \underline{0.9409 ± 0.0025} & \textbf{0.9550 ± 0.0013} \\
& & ICEWS1819 & 0.9333 ± 0.0026 & 0.9134 ± 0.0041 & \underline{0.9716 ± 0.0033}& 0.9631 ± 0.0034& \textbf{0.9789 ± 0.0022}& 0.9625 ± 0.0030& 0.9688 ± 0.0018 & 0.9634 ± 0.0010\\
& & Googlemap CT & \textit{OOM} & \textit{OOM} & \underline{0.8750 ± 0.0015} & 0.8012 ± 0.0021 & 0.7936 ± 0.0009 & 0.7633 ± 0.0013 & 0.7735 ± 0.0031 & \textbf{0.8890 ± 0.1134}\\
& & GDELT & 0.9019 ± 0.0023 &  0.8928 ± 0.0011 & 0.9023 ± 0.0010 & 0.8986 ± 0.0077 & 0.9151 ± 0.0045 & 0.8925 ± 0.0048 & \underline{0.9263 ± 0.0009} & \textbf{0.9369 ± 0.0056}\\ 
\hline 
 \end{tabular}}
\caption{\selectfont Performance comparison of \textit{ROC-AUC} and \textit{AP} for \textbf{future link prediction} for transductive  (\textit{tr.})  and inductive (\textit{in.}) setting. (\textit{OOM} means Out-Of-Memory.)}
\vspace*{-1em}
\label{tab:flp_benchmark}
\end{table*}

\begin{table*}[!ht]
\centering
\scalebox{0.65}{
\begin{tabular}{c|c|c|c|c|c|c|c|c|c}
\toprule \textbf{Datasets} & \textbf{Metrics} & \textbf{JODIE} & \textbf{DyRep} & \textbf{TGAT} & \textbf{CAWN} & \textbf{TCL} & \textbf{GraphMixer} & \textbf{DyGFormer} & \textbf{\proj} \\
\bottomrule \multirow{3}{*}{ Enron } & Precision & \underline{0.6568 ± 0.0043} & 0.6635 ± 0.0052 & 0.6148 ± 0.0012 & 0.6076 ± 0.0070 & 0.5530 ± 0.0079 & 0.6313 ± 0.0024 & 0.6601 ± 0.0067 & \textbf{0.6685 ± 0.0013} \\
 & Recall &\underline{ 0.6472 ± 0.0039 }& 0.6390 ± 0.0089 & 0.5530 ± 0.0001 & 0.5783 ± 0.0094 & 0.5394 ± 0.0061 & 0.5735 ± 0.0015 & 0.5802 ± 0.0071 & \textbf{0.6567 ± 0.0013} \\
 & F1 & \underline{0.6478 ± 0.0065} & 0.6432 ± 0.0062 & 0.5519 ± 0.0028 & 0.5685 ± 0.0132 & 0.5177 ± 0.0044 & 0.5507 ± 0.0019 & 0.5604 ± 0.0063 & \textbf{0.6675 ± 0.0014} \\
\hline \multirow{3}{*}{ GDELT } & Precision & 0.1361 ± 0.0036 & 0.1451 ± 0.0071 & 0.1241 ± 0.0056 & \underline{0.1781 ± 0.0011} & 0.1229 ± 0.0021 & 0.1293 ± 0.0026 & 0.1775 ± 0.0041 &  \textbf{0.2139 ± 0.0012}\\
 & Recall & 0.1338 ± 0.0013 & 0.1365 ± 0.0013 & 0.1321 ± 0.0012 & 0.1545 ± 0.0001 & 0.1235 ± 0.0047 & 0.1320 ± 0.0008 & \underline{0.1580 ± 0.0052} & \textbf{0.2159 ± 0.0016} \\
 & F1 & 0.0992 ± 0.0009 & 0.1039 ± 0.0012 & 0.0967 ± 0.0010 & \underline{0.1340 ± 0.0012} & 0.0987 ± 0.0051 & 0.1014 ± 0.0017 & 0.1291 ± 0.0068 & \textbf{0.2234 ± 0.0023}\\
\hline \multirow{3}{*}{ ICEWS1819 } & Precision & 0.3106 ± 0.0023 & 0.3270 ± 0.0025 & 0.3013 ± 0.0007 & \underline{0.3451 ± 0.0023} & 0.3212 ± 0.0096 & 0.2999 ± 0.0022 & 0.3297 ± 0.0034 & \textbf{0.3743 ± 0.0043}\\
 & \begin{tabular}{l} 
Recall
\end{tabular} & 0.3494 ± 0.0018 & 0.3636 ± 0.0020 & 0.3512 ± 0.0006 & \underline{0.3676 ± 0.0034} & 0.3517 ± 0.0009 & 0.3502 ± 0.0001 & 0.3632 ± 0.0026 & \textbf{0.3823 ± 0.0013}\\
 & F1 & 0.2965 ± 0.0008 & 0.3097 ± 0.0006 & 0.2908 ± 0.0008 & \underline{0.3156 ± 0.0057} & 0.2939 ± 0.0022 & 0.2903 ± 0.0008 & 0.3079 ± 0.0027 & \textbf{0.3243 ± 0.0013}\\
\hline \multirow{3}{*}{ Googlemap CT } & Precision & 0.6163 ± 0.0032 & 0.6073 ± 0.0019 & 0.6160 ± 0.0001 & 0.6166 ± 0.0023 & \underline{0.6213 ± 0.0087} & 0.6171 ± 0.0020 & 0.6166 ± 0.0003 & \textbf{0.6305 ± 0.0057}\\
 & Recall & 0.6871 ± 0.0002 & 0.6827 ± 0.0006 & 0.6862 ± 0.0002 & 0.6870 ± 0.0001 & 0.6875 ± 0.0001 & 0.6872 ± 0.0003 & \underline{0.6877 ± 0.0002} & \textbf{0.6905 ± 0.0002}\\
 & F1 & 0.6189 ± 0.0016 & 0.6134 ± 0.0006 & 0.6225 ± 0.0015 & 0.6187 ± 0.0003 & \underline{0.6230 ± 0.0003} & 0.6185 ± 0.0005 & 0.6196 ± 0.0008 & \textbf{0.6354 ± 0.0012}\\
\hline \multirow{3}{*}{ Stack elec } & Precision & \textit{OOM} & \textit{OOM} & 0.6265 ± 0.0046 & 0.6167 ± 0.0094 & \underline{0.6325 ± 0.0023 }& 0.6074 ± 0.0039 & 0.6026 ± 0.0471 & \textbf{0.6565 ± 0.0056}\\
 & Recall & \textit{OOM} & \textit{OOM} & 0.7205 ± 0.0094 & 0.6313 ± 0.0462 & \underline{0.7474 ± 0.0004} & 0.7412 ± 0.0061 & 0.5891 ± 0.2747 & \textbf{0.7556 ± 0.0056}\\
 & F1 & \textit{OOM} & \textit{OOM} & \underline{0.6496 ± 0.0032} & 0.6209 ± 0.0216 & 0.6420 ± 0.0003 & 0.6412 ± 0.0005 & 0.4860 ± 0.2686 & \textbf{0.6796 ± 0.0024}\\
\hline \multirow{3}{*}{ Stack ubuntu } & Precision & \textit{OOM} & \textit{OOM} & 0.6858 ± 0.0047 & \underline{0.6921 ± 0.0040} & 0.6915 ± 0.0118 & \textbf{0.6930 ± 0.0028} & 0.6789 ± 0.0490 & 0.6824 ± 0.0054\\
 & Recall & \textit{OOM} & \textit{OOM} & \textbf{0.7921 ± 0.0012} & 0.5650 ± 0.1015 & 0.7880 ± 0.0026 & \underline{0.7902 ± 0.0130} & 0.7494 ± 0.0991 & 0.7554 ± 0.1224\\
 & F1 & \textit{OOM} & \textit{OOM} & 0.7201 ± 0.0013 & 0.6002 ± 0.0738 & \textbf{0.7219 ± 0.0046} & \underline{0.7214 ± 0.0014} & 0.7033 ± 0.0294 & 0.7156 ± 0.0056\\  \hline
\end{tabular}}
\caption{\selectfont Performance of dynamic graph learning methods on \textbf{edge classification} task. (\textit{OOM} means Out-Of-Memory.)}
\vspace*{-1em}
\label{tab:edge_classification_benchmark}
\end{table*}

\section{Experiments}

To comprehensively verify the effectiveness of \proj, we conduct extensive experiments and attempt to answer the following questions:

\begin{itemize}
    \item How well \proj, utilizing knowledge distillation technique from teacher LLM to student GNN model, comprehend the semantics of link creation and edge types for the future link prediction and edge classification tasks?
    \item How does knowledge distillation impacts the performance of \proj in future link prediction task under both transductive and inductive setting?
    \item What is the impact of the knowledge distillation loss on future link prediction and edge classification task?
\end{itemize}
\subsection{Experimental Setup}
\textbf{Dataset and Baseline.} To validate the efficacy of \proj on future link prediction and edge classification task, we compare with the performance of \proj with seven baseline from the DTGB~\cite{zhang2024dtgb} paper for six Dynamic Text-Attributed Graphs.  We present the statistics of the dynamic text-attributed graphs used in the benchmark in Table~\ref{tab:DyTAG_datasets}.  Among the tested baseline methods, we include the RNN-based approach JODIE~\cite{kumar2019predicting}, DyReP~\cite{trivedi2019dyrep}, self-attention based approach TGAT~\cite{xu2020inductive}, causal anonymous random-walk based approach CAWN~\cite{wang2021inductive}, transformer based approach TCL~\cite{wang2021tcl}, fixed time-encoding and summarization based approach GraphMixer~\cite{cong2023we}, and patching with transformer based DyGFormer~\cite{yu2023towards}. A detailed description of the datasets is provided in the Appendix.

\textbf{Experimental Settings.} In our experimental setting, we first obtain the node and edge embeddings from a comparatively lightweight language model BERT~\cite{devlin2018bert}. Also, for each edge we add a zero-vector of shape equal to the total number of timestamps. We set it as $1$ from the timestamp when an edge has occurred until the end to denote the existence of an edge, and use $-1$ to mask the future part. We do not consider edge drop which can be denoted using $0$. 
We concatenate the BERT embedding with the temporal edge encoding. For obtaining the spatio-temporal embedding, we use GATConv~\cite{kipf2016semi} and TransformerConv~\cite{shi2020masked} from Pytorch geometric package\footnote{https://pytorch-geometric.readthedocs.io/en/latest/}
which utilizes edge features during graph convolution to produce node features. 
For all datasets, we split the data into training, validation, and test sets using a  \textit{70/15/15} ratio based on the timestamps of the edges. Additionally, during training we sample an equal number of fake edges not originally present in the dataset for the future link prediction task. This helps train the student model to distinguish between true and fake edges through the knowledge distillation process. The temporal encoding for the fake edges is kept as a zero vector. We run each experiment on every dataset five times for $50$ epochs and report the mean and standard deviation for difference metrics.

To obtain textual representations from the text prompt efficiently, we use superfastllm~\footnote{\url{https://huggingface.co/power-greg/super-fast-llm}} from Huggingface~\cite{wolf2019huggingface} as our LLM model. We compute the BERT embeddings and the textual edge embedding as a pre-processing step for training of each epoch. 
For the future link prediction task, we adopt the AUC-ROC metric to compare the performance between ~\proj and the baseline models. For the edge classification task, we report precision, recall, and F1 score to compare the performance difference.

\textbf{Hyperparameter Tuning.} To tun
the performance of our models, we perform grid search of the hyperparameters of \proj as follows: 1) Dimensionalities of student and teacher models $d_{student}, d_{teacher} \in \{16, 32, 64, 128, 256\}$; 2) Number of MLP layers of $\phi_{mlp\_student}$ and $\phi_{mlp\_teacher}$ $\in$ $ \{1,2,3\}$; 3) Number of layers in message passing GNN $\psi$ $\in \{1,2,3\}$; 4) Future Link Prediction, Edge Classification, Knowledge Distillation hyperparameter $\lambda_{flp}, \lambda_{ec}, \lambda_{kd} \in \{0.25, 0.5, 0.75, 1.0, 1.25, 1.5, 1.75, 2.0\}$; 5) GNN Type $\in \{\text{GATConv, TransformerConv}\}$; 6) Batch Size $ \in \{32, 64, 128, 256, 512, 1024, 2048\}$; 7) learning rate $\in \{0.01, 0.001, 0.0001\}$. Also, we choose Adam optimizer~\cite{kingma2014adam} to update the parameters of the student model using the task-specific and knowledge distillation loss.

\textbf{Hardware.} All experiments are conducted on a Linux server equipped with a 2.20 GHz Intel Xeon E5-2650 v4 processor and four NVIDIA Tesla V100 GPUs, each with 32 GB of VRAM.

\subsection{Results}

\textbf{Future Link Prediction and Edge Classfication Results.} We present the performance comparsion of ~\proj with baseline approaches for the future link prediction task in the Table ~\ref{tab:flp_benchmark} and for the edge classification task in the Table~\ref{tab:edge_classification_benchmark}. From Table~\ref{tab:flp_benchmark}, we observe that ~\proj outperforms the baseline approaches in the future link prediction task for all four datasets in the transductive setting. From this we can conclude that ~\projshort, with the temporal encoding and knowledge distillation loss, effectively captures both the temporal dynamics and the semantics of edge creation for future link prediction in the transductive setting. However, for inductive future link prediction, we observe that the performance of \proj on the ICEWS1819 is not superior. Note that ICEWS1819 dataset represents the edges between political entities where the timestamp has much sparser granularity, which makes it challenging to grasp the semantics behind edge creation when new nodes appear in the test timestamps. This may explain \projshort's performance degradation compared to state-of-the-art approaches in inductive link prediction on the ICEWS1819 dataset. 
For the edge classification task, we present the performance comparison on precision, recall, and F1 score in the Table~\ref{tab:edge_classification_benchmark}. We observe that ~\proj can outperform the baseline approaches in all the datasets in all three metrics except for the Stack ubuntu dataset. In stack ubuntu, the text information of the answer edges include code as well as plain text between users and questions while the edge label is whether the answer is useful or not useful. The mixture of code and natural language might make it difficult to understand the semantics and thus the performance of \proj for edge classification is not superior compared to baseline models.

\begin{figure}[!t]
\scalebox{0.9}{\includegraphics[width=0.5\textwidth, left]{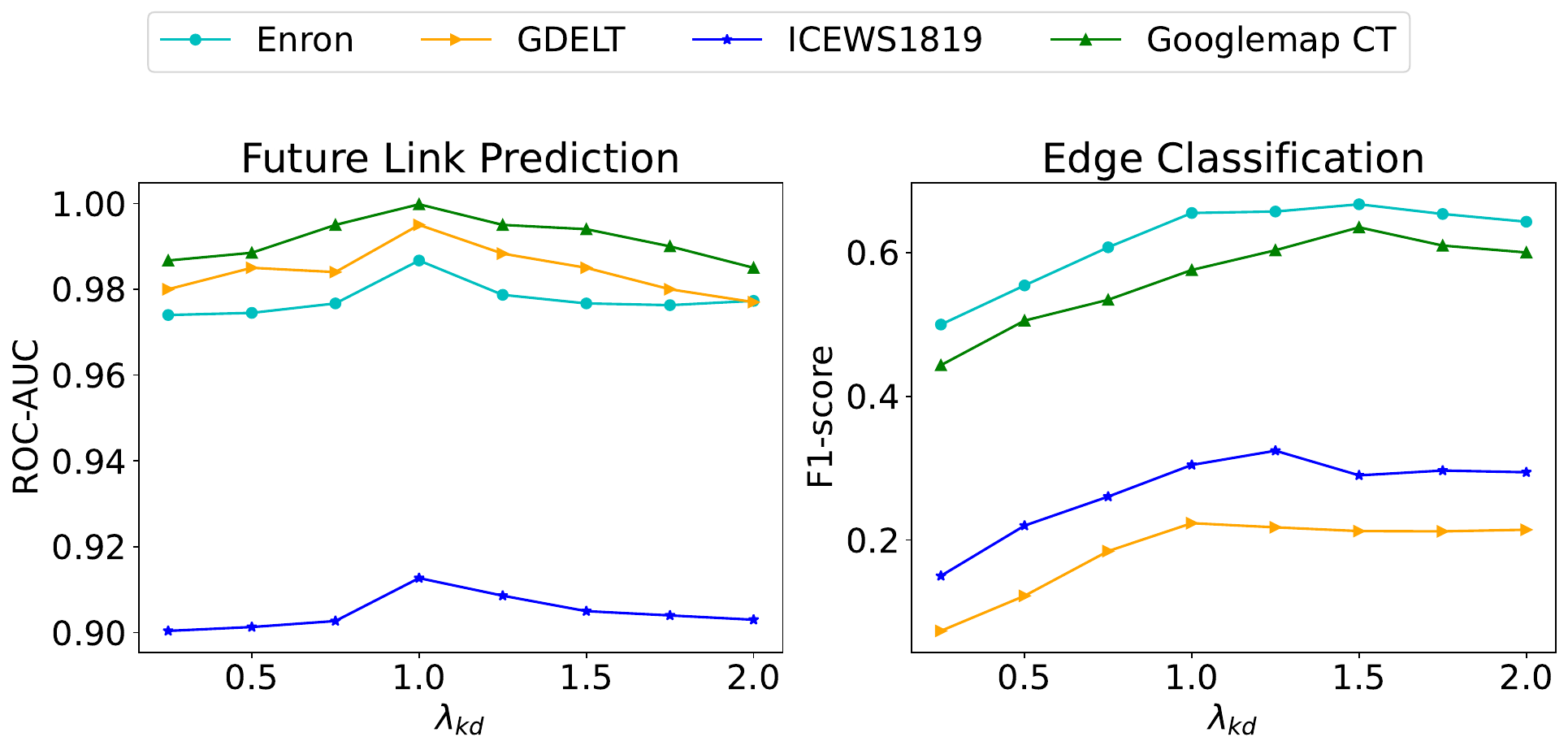}}
\caption{\fontsize{10}{10} \selectfont The impact of $\lambda_{kd}$ for \textbf{future link prediction} and \textbf{edge classification} tasks.}
\vspace*{2em}
\label{fig:kd_flp_ec}
\end{figure}

\textbf{Impact of Knowledge distillation for future link prediction and edge classification.} To understand how knowledge distillation process impact future link prediction and edge classification tasks, we vary the hyperparameter $\lambda_{kd} \in \{0.25, 0.5, 0.75, 1.0, 1.25, 1.5, 1.75, 2.0\}$ to analyze the impact on both tasks and present the results in Figure~\ref{fig:kd_flp_ec}. Intuitively, the creation of a future edge should primarily rely on the text information of neighboring nodes, with structural and temporal information also playing a significant role. On the other hand, the edge classification should mostly depend on the semantic meaning of the text description of the edge to determine the edge label. From Figure~\ref{fig:kd_flp_ec}, we can observe a clear trend on the edge classification task, as $\lambda_{kd}$ increases, the performance improves across all datasets up to a certain point and then stabilizes.
For future link prediction task, the performance change is less pronounced when the impact of knowledge distillation loss increases. 
Therefore, semantic information has a greater impact on \proj's edge classification performance than on future link prediction. From this ablation study of knowledge distillation, we conclude that semantic information is more critical for edge classification task than the future link prediction task in Dynamic Text-Attributed Graphs (DyTAGs). 

\begin{figure}[!t]
\scalebox{0.8}{
\includegraphics[width=0.6\textwidth,left]{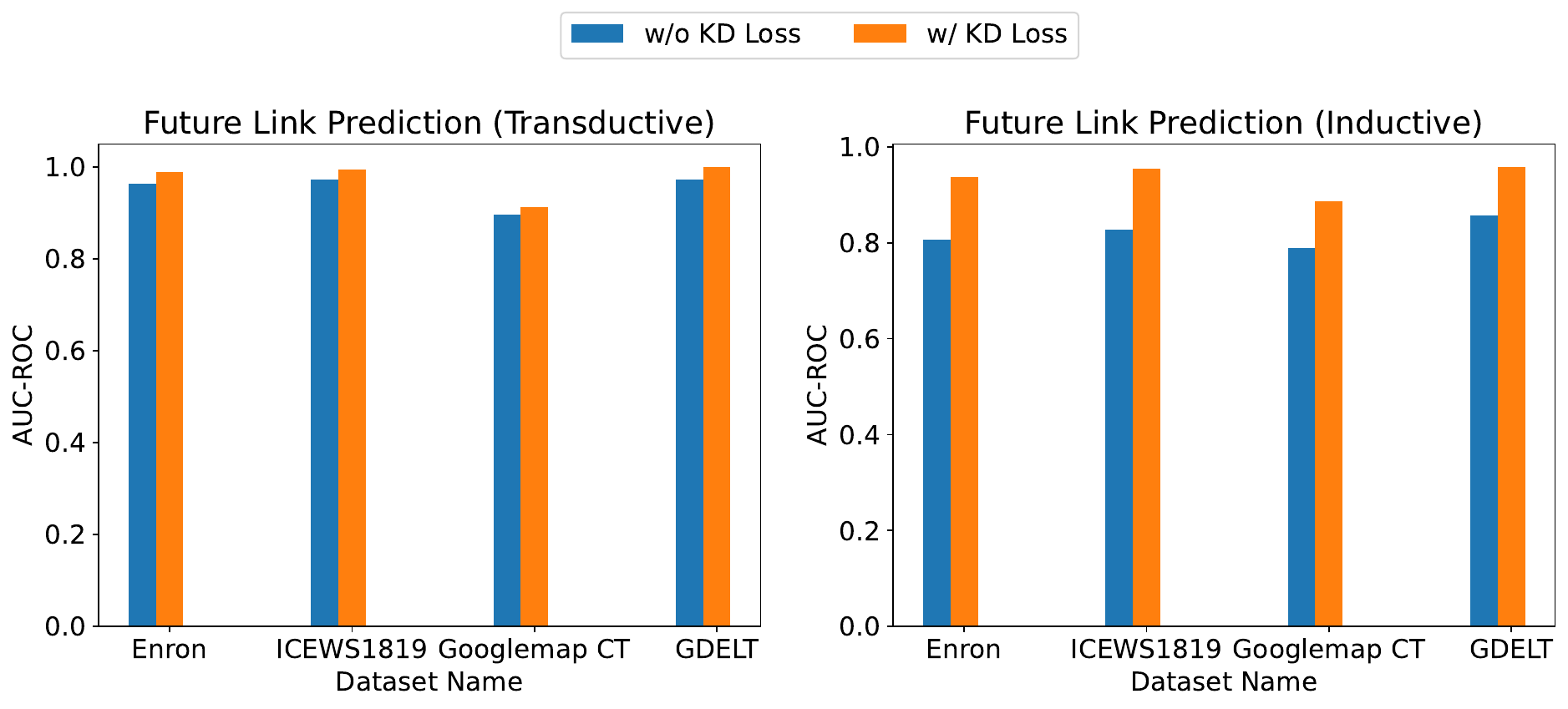}}
\caption{\fontsize{10}{10} \selectfont The impact of knowledge distillation on the \textbf{future link prediction} task for the inductive setting (right) and the transductive setting (left).}
\vspace*{2em}
\label{fig:kd_loss_ablation}
\end{figure}

\textbf{Knowledge Distillation for transductive and inductive link prediction.} To analyze the impact of semantic information from textual edge representation on the future link prediction task for transductive and inductive settings, we conduct experiments with and without the knowledge distillation loss. From Figure~\ref{fig:kd_loss_ablation}, for different datasets, the performance difference with knowledge distillation loss is more significant for the inductive setting than for the transductive setting. 
We observe that adding knowledge distillation impacts the performance of future link prediction task more on the inductive setting. In inductive link prediction, new nodes are introduced during inference, and knowledge distillation enhances the student GNN model's ability to understand the semantic context of edge creation. In the inductive setting, where new nodes are introduced during inference, the impact of knowledge distillation is more pronounced compared to the transductive setting, where the same nodes are present during both training and inference. Therefore, the ability to understand the semantics of edge creation with new nodes lead to better future link predictions for \proj.




\section{Conclusion}
We propose a simple yet efficient framework that encodes time in edges to integrate temporal and structural information together into the spatio-temporal representation. For the text component, we use knowledge distillation from LLM-driven textual representations of edges in dynamic text-attributed graphs. We show that  distilling knowledge from text-based edge representations of the teacher LLM to the student GNN enables us to achieve state-of-the-art performance on benchmark datasets of dynamic text-attributed graphs for both future link prediction and edge classification tasks. Future directions of this work may possibly include exploration of very large dynamic text attributed graphs.


\clearpage

\bibliography{aaai25}

\clearpage

\section{Supplementary Materials}

\subsection{Notations} We present the notations used to describe our model \proj in Table ~\ref{tab:notation}.

\begin{table*}[!t]
\centering
\begin{tabular}{c|c}\hline 
 \textbf{Notation} & \textbf{Description}\\ \hline
$\mathcal{G}$ & Dynamic Text Attributed Graph \\
$\mathcal{V}, \mathcal{E}$ & Set of Node and Edges \\
$e_{uv}$ & Edge between node $u$ and node $v$ \\ 
$r_{uv}$ & Relation between node $u$ and $v$ \\
$t_{uv}$ & Timestamp of the edge between node $u$ and $v$\\
$l_{uv}$ & Label of the edge between node $u$ and node $v$ \\
$u_{text}, v_{text}$ & Text Description of node $u$ and $v$ \\
$r_{text}$ & Text description of an edge relation $r$\\
$\mathcal{T}$ & Set of timestamps\\
$\mathcal{L}$ & Set of edge labels\\
$\mathcal{N}_u$ & Neighbor set of node $u$\\
$\mathcal{G}_T = (\mathcal{V}_T, \mathcal{E}_T)$ & Dynamic Graph until timestmp $T$\\
$k$ & Future edge timestamps to be predicted\\
$\tau$ & Temporal Edge Encoding\\
$h_u$, $h_v$ & Embedding for nodes $u$ and $v$ (Student Model)\\
$h_{uv}$ & Embedding for an edge $e_{uv}$ (Student Model)\\
$\Hat{h}_{uv}$ & Edge embedding concatenated with temporal encoding\\
$\phi_{mlp\_student}$ & MLP to transform the frozen BERT embedding to node representation (Student Model)\\
$\psi$ & Message passing GNN (Student Model)\\
$h_{spatio\_temporal}$ & Spatio-temporal Representation of an edge (Student Model)\\
$d_{BERT}$ & Dimensionality of the BERT embeddings \\
$d_{student}$ & Embedding for the spatio-temporal representation from student model\\
$\phi_{flp}$ & MLP for future link prediction (Student Model)\\
$\phi_{ec}$ & MLP for edge classification (Student Model)\\
$\Hat{\tau}$ & Student Model predicted Temporal Encoding\\
$\Hat{l}$ & Student Model predicted label vector\\
$h_{\mathcal{N}_u}, h_{\mathcal{N}_v}$ & Neighborhood Textual Embedding of node $u$ and node $v$\\
$\phi_{mlp\_teacher}$ & MLP to transform the neighborhood textual representation\\
$d_{LLM}$ & Dimensionality of the LLM embeddings \\
$d_{teacher}$ & Embedding for the textual representation from teacher model\\
$\Tilde{h}_{uv}$ & LLM transformed output of link prompt of an edge $e_{uv}$ \\
$h_{textual}$ & Textual edge representation (Teacher Model)\\
$\Bar{h}_{spatio\_temporal}$ & Normalized Spatio-Temporal Edge Representation (Teacher Model)\\
$\Bar{h}_{textual}$ & Normalized Textual Edge Representation (Teacher Model)\\
$L_{flp}$ & Future Link Prediction loss\\
$L_{flp}$ & Edge Classification loss\\
$L_{kd}$ & Knowledge Distillation loss\\
$BERT(.)$ & Function call to BERT model with Text to obtain embedding \\
$LLM(.)$ & Function call to any LLM with Text to obtain embedding \\ \hline
\end{tabular}
\caption{\small Notations used in \proj}
\label{tab:notation}
\end{table*}

\subsection{Psedocode}

\begin{algorithm}[!ht]
\caption{\small \textbf{\proj}: LLM-driven Knowledge Distillation for Dynamic Text-Attributed Graph with temporal encoding}
\label{alg:LLMKD4DyTAG}
\begin{algorithmic}[1]
\State{\textbf{Preprocessing}:}
\Function{$neighbor\_prompt$}{$e_{uv}$}:
\State{\textit{prompt = ``Entity $u_{text}$ is connected to entity $v_{text}$ via relation $r_{uv}$ at timestamp $t_{uv}$ with label $l_{uv}$"}}
\State{\textit{return prompt}}
\EndFunction
\Function{$link\_prompt$}{$e_{uv}$}:
\State{\textit{prompt = ``There is \textbf{an/no} edge between entity $u_{text}$ and entity $v_{text}$ via relation $r_{uv}$ at timestamp $t_{uv}$ with label $l_{uv}$"}}
\State{\textit{return prompt}}
\EndFunction
\Function{$compute\_textual\_representation$}{$V,E$}
\For{ each node $u \in V$}
\State{$h_u = BERT(u_{text}) $}
\State{\small $h_{\mathcal{N}(u)} = \sum_{w \in \mathcal{N}(u)} LLM(neighbor\_prompt(e_{uw}))$}
\EndFor
\For {each edge $e_{uv} \in E$} \Comment{ $e_{uv} = (r_{uv}, t_{uv}, l_{uv})$}
\State{$h_{uv} = BERT(r_{uv})$}
\State{$\Tilde{h}_{uv} = LLM(link\_prompt(e_{uv}))$}
\State{$h_{textual}  = h_{\mathcal{N}(u)} + h_{\mathcal{N}(v)} + \Tilde{h}_{uv}  $}
\EndFor
\EndFunction \\ \hrulefill
\State {\textbf{\proj:}}
\State {\textbf{Input:} \textbf{DyTAG} $G = (V,E)$ with timestamp and label set $\mathcal{T,L}$ for each edge} 
\State{\textit{compute\_textual\_representation (V,E)}}
\For {each edge $e_{uv} \in E$} \Comment{ $e_{uv} = (r_{uv}, t_{uv}, l_{uv})$}
\State{$ \tau[:t_{uv}] = 0$,  $  \tau[t_{uv}:T+1] = 1$}
\State{$\tau[T:T+k+1] = -1$, $\tau \in \mathbb{R}^{T+k}$}
\State{$\Hat{h}_{uv} = [h_{uv} || \tau]$} \scalebox{0.77}{\Comment{Temporal Encoding : Concatenate $h_{uv}$ and $\tau$}}
\State{$\Bar{h}_{u} = \text{UPDATE}(h_u, \text{AGG}(\{(h_w,\Hat{h}_{uw}): w \in \mathcal{N}_u\}))$}
\State{$\Bar{h}_{v} = \text{UPDATE}(h_v, \text{AGG}(\{(h_w,\Hat{h}_{vw}): w \in \mathcal{N}_v\}))$}
\State{$h_{spatio\_temporal} = \Bar{h}_{u} \odot \Bar{h}_{v}
$}
\State{$\Hat{\tau} = \sigma(\phi_{flp}(h_{spatio\_temporal}))$}
\State{$\Hat{l}_{uv} = \sigma(\phi_{ec}(h_{spatio\_temporal}))$}
\State{$\bar{h}_{textual} = \frac{h_{textual}}{||h_{textual}||}$}
\State{$\bar{h}_{spatio\_temporal} = \frac{h_{spatio\_temporal}}{||h_{spatio\_temporal}||}$}
\State{$\mathcal{L}_{flp} = - \sum_{i=0}^{T+k} \tau_i \log( \Hat{\tau_i})$}
\State{$\mathcal{L}_{ec} = - \sum_{i=0}^{|\mathcal{L}|} l_i \log( \hat{l}_i)$}
\State{$\mathcal{L}_{kd} = \textit{e}^{-(\bar{h}_{textual}^\top \bar{h}_{spatio\_temporal})}$}
\State{$\mathcal{L}_{flp\_with\_kd} = \lambda_{flp} \mathcal{L}_{flp} + \lambda_{kd} \mathcal{L}_{kd}  $} \scalebox{0.65}{\Comment{ Future Link Prediction}}
\State{$\mathcal{L}_{ec\_with\_kd} =  \lambda_{ec} \mathcal{L}_{ec} + \lambda_{kd} \mathcal{L}_{kd}  $} \scalebox{0.8}{\Comment{\small Edge Classification}}
\EndFor
\end{algorithmic}
\end{algorithm}

\subsection{Time Complexity of \proj}

As a pre-processing step, we use BERT encoder to obtain the embeddings for the node and edge text. Let us consider the number of layers in the BERT encoder is $L_{BERT}$, the input text length is $n_{text}$ and the hidden dimension of BERT is $d_{BERT}$.

Then, to compute the self-attention part of BERT, the time complexity will be $O(n_{text}^2 . d_{BERT})$ since each token depends on every other token while the feed-forward layers will take $O(n_{text} . d_{BERT}^2)$ For each layer of BERT, the total time complexity will be $O (n_{text}^2 . d_{BERT} + n_{text} . d_{BERT}^2)$ and for $L_{BERT}$ layers it will be $O (L_{BERT} . (n_{text}^2 . d + n_{text} . d_{BERT}^2))$.
Also, since we use text prompts for each edge in the teacher LLM to obtain teacher embeddings as a pre-processing step the time complexity will be similar i.e.  $O (|E| L_{LLM} . (n_{text}^2 . d_{teacher} + n_{text} . d_{teacher}^2))$ where $d_{teacher}$ is the dimension of teacher LLM and $L_{LLM}$ is the dimension of the teacher LLM and $|E|$ is the total number of edges in the dynamic text attributed graph.

To obtain the edge encoding the time complexity will be $O(|E| . |T|)$ where $|E|$ denotes the total edges in the dynamic text-attributed graph while $|T|$ denotes the total number of timestamps.

We run student GNN to obtain the representation for the dynamic text-attributed graphs which can be parallelized and take the time complexity $O(L_{GNN} (n^2 + |E|d_{student}))$ where $L_{GNN}$ is the number of layers in the GNN, n is the number of nodes in the dynamic text attributed graph and $|E|$ is the total number of edges in the dynamic text attributed graph and $d_{student}$ is the dimension of the student GNN model.
\subsection{Related Works}

\textbf{Dynamic Graphs.} JODIE~\cite{kumar2019predicting} uses coupled recurrent neural networks to forecast embedding trajectories for entities, enabling predictions of their temporal interactions. DyRep~\cite{trivedi2019dyrep} integrates recurrent node state updates with a deep temporal point process and temporal-attentive network to model evolving graph dynamics nonlinearly. TGAT~\cite{xu2020inductive} utilizes self-attention for aggregating temporal-topological neighborhood features and captures temporal patterns with a functional time encoding method based on Bochner's theorem. CAWN~\cite{wang2021inductive} employs an anonymization strategy using sampled walks to explore network causality and generate node identities, which are encoded and aggregated using a neural network model to produce the final node representation. TCL~\cite{wang2021tcl} utilizes a two-stream encoder for temporal neighborhoods of interaction nodes, integrating a graph-topology-aware Transformer with cross-attention to learn node representations considering both temporal and topological information. GraphMixer~\cite{cong2023we} demonstrates the effectiveness of fixed-time encoding for dynamic interactions using a simple architecture with components for link summarization, node summarization, and link prediction. DyGFormer~\cite{yu2023towards} learns node representations from historical first-hop interactions using a neighbor co-occurrence encoding scheme and a patching technique to effectively leverage longer historical sequences. To include textual information into the nodes and edges of dynamic graphs, a recent benchmark, DTGB~\cite{zhang2024dtgb} formally defines Dynamic Text-Attributed Graph and performs baseline comparisons on future link prediction and edge classification tasks.

\textbf{LLMs for Temporal Data.} STD-LLM~\cite{huang2024std} proposes spatial and temporal tokens and hypergraph learning module to effectively capture non-pairwise and higher-order spatial-temporal correlations, which also helps to understand the capabilities of large language models (LLMs) for spatio-temporal forecasting and imputation tasks.  STLLM~\cite{zhang2023spatio} proposes to use spatio-temporal prompts to obtain representations from LLMs, which is further aligned with GNNs representations using the InfoNCE loss. In another work~\cite{liu2024spatial}, spatio-temporal embeddings are input into both frozen and unfrozen transformer blocks to produce representations for traffic prediction. STGLLM~\cite{liu2024can} proposes a tokenizer and a LLM-based adapter for performing traffic prediction.

\textbf{Knowledge Distillation involving LLMs and GNNs.} There are several recent works focus on using knowledge distillation to transfer the text processing capabilities of LLMs to lightweight models~\cite{hsieh2023distilling} like GNNs for text-attributed graphs. 
LinguGKD~\cite{xu2024llm} proposes a model for distilling knowledge from $k$-hop prompt representations in an LLM to $k$-hop GNN representations. However, their approach describe the entire $k$-hop neighborhood which is not well-suited for graphs with long text in nodes and edges. 
LLM4TAG~\cite{pan2024distilling} proposes to learn text-attributed graph by knowledge distillation from LLMs to GNNs. Another recent work~\cite{luo2024chain} utilizes LLMs for temporal knowledge graph completion by leveraging LLMs' capabilities of reasoning with particular attention to reverse logic.
GAugLLM~\cite{fang2024gaugllm} proposes an approach for improving the contrastive learning for Text-Attributed Graph by mixture of prompt experts. AnomalyLLM~\cite{liu2024large} proposes a method for time series anomaly detection by knowledge distillation. UniGLM~\cite{fang2024uniglm} proposes an unified graph language model that generalizes both in-domain and cross-domain TAGs. GALLON~\cite{xu2024llm} proposes a multi-modal knowledge distillation strategy to transfer the text and structure processing capability of LLMs and GNNs into MLPs for molecular property prediction.\\

\subsection{Dataset Description}

\subsubsection{Data Format:} For each of the Dynamic Text-Attributed Graph (DyTAG) dataset, we have three different files \textit{edge\_list.csv}, \textit{entity\_text.csv}, \textit{relation\_text.csv}.
The \textit{edge\_list.csv} file contains the dynamicallly evolving grpah structure, where each line describe the source node id, the destination node id, the relation id, timestamp and label of the edge. The \textit{entity\_text.csv} and \textit{relation\_text.csv} contains the mapping of node id and relation id to the rich text description. We download the datasets from the github repository~\footnote{https://github.com/zjs123/DTGB} of the benchmark paper DTGB~\cite{zhang2024dtgb}. We present a short description of each of the six DyTAG dataset below.

\begin{itemize}
    \item \textbf{Enron}~\footnote{https://www.cs.cmu.edu/~enron/}: This dataset is derived from the email interactions between the employees of  ENRON energy corporation from 1999 to 2002. The nodes contain the employee profiles and their position and the edges include the truncated emails between them without non-English statement, abnormal symbols, tables from the raw text. The edge categories are the type of the emails e.g. calendar, notes, deal communication etc. The edges of this dataset are ordered by the sending time of the email.
    \item \textbf{GDELT}~\footnote{https://www.gdeltproject.org/}: This dataset is based on the political behavior across all countries in the world, derived by the Global Database of Events, Language, and Tone project. The nodes denote the political entities e.g. ``United States", ``Donald Trump" whereas the edges denote the relationship between these nodes e.g. ``born in", ``president of". Edge categories are based on the political relationship/type of behavior and the edges are ordered based on the timestamp of the occuring events.
    \item \textbf{ICEWS1819}~\footnote{https://dataverse.harvard.edu/dataverse/icews}: This is another dynamic text-attributed graph dataset based on political events from 2018-01-01 to 2019-12-31 derived by the  Integrated Crisis Early Warning System project. The name, sector and nationality of each political entity is used as the text attribute and the edges denote the political relationship. The edge categories represents the political relationship/behavior and the edges are sorted based on the timestamp of the occurring events. The difference between GDELT and ICEWS1819 is that GDELT describes the event in a more fine-grained way (15 minutes of time interval) whereas the events of  ICEWS1819 are more sparse grained (24 hours of time interval). The number of nodes of ICEWS1819 is 4 times more than the GDELT dataset, therefore represents more sparse scenario. 
    \item \textbf{Stack elec}~\footnote{https://archive.org/details/stackexchange}: Stack exchange data contains anonymized stack exchange data with question, answer, comments, tags etc. The nodes denote the question and user whereas the edges denote the answer and comments between the user node and question node, describing an dynamic bipartite graph which allow multi-round dialogue between users and questions. The stack elec dataset contains stack exchange data where the questions are related with electronic techniques and their corresponding answers and comments.  For user nodes the self-introduction and the name of the technical areas of users expertise are used as the text whereas for the question nodes the title and body of each question are used as the text attributes. For the edges the text of the answer is used as the text attribute.Two types of edge categories are defined based on the answers, \textit{useful} if the number of votes on the answer is greater than 1 else \textit{useless}. The edges are ordered based on the answering timestamp from users. 
    \item \textbf{Stack ubuntu}~\footnote{https://archive.org/details/stackexchange}: Stack ubuntu dataset is another stack exchange dataset containing the questions related to ubuntu system. The answers contains mixture of codes and natural language which makes the understanding of semantic context of interactions more challenging.  
    \item \textbf{Googlemap CT}~\footnote{https://datarepo.eng.ucsd.edu/mcauley\_group/gdrive/googlelocal/}: This dataset is a review dataset extracted from the Google Local Data project, containing review information on google map between users and business entities of the Connecticut state up to September 2021. Nodes are users and business entities whereas the edges are reviews of from user nodes to business nodes. Only the business nodes contains the text descriptions which is the name, address, category, and self-introduction of the business entity. The edge text attributes are the raw text of user reviews without the emojis and meaningless characters. The edge categories are review ranting ranging from 1 to 5. The edges in the dataset are ordered based on the timestamp of the review.
\end{itemize}

\begin{table*}[!t]
\centering
\begin{tabular}{c|c|c|c}
\hline
\textbf{Dataset}                       & \textbf{Metrics} & \textbf{w/o Temporal Encoding} & \textbf{with Temporal Encoding} \\ \hline
\multirow{3}{*}{\textbf{Enron}}        & Precision        & 0.6456 ± 0.0076                & \textbf{0.6685 ± 0.0013}        \\ \cline{2-4} 
                                       & Recall           & 0.6212 ± 0.0056                & \textbf{0.6567 ± 0.0013}        \\ \cline{2-4} 
                                       & F1               & 0.6345 ± 0.0073                & \textbf{0.6675 ± 0.0014}        \\ \hline
\multirow{3}{*}{\textbf{GDELT}}        & Precision        & 0.1733 ± 0.0034                & \textbf{0.2139 ± 0.0012}        \\ \cline{2-4} 
                                       & Recall           & 0.1843 ± 0.0076                & \textbf{0.2159 ± 0.0016}        \\ \cline{2-4} 
                                       & F1               & 0.1834 ± 0.0123                & \textbf{0.2234 ± 0.0023}        \\ \hline
\multirow{3}{*}{\textbf{ICEWS1819}}    & Precision        & 0.3212 ± 0.0056                & \textbf{0.3743 ± 0.0043}        \\ \cline{2-4} 
                                       & Recall           & 0.3357 ± 0.0076                & \textbf{0.3823 ± 0.0013}        \\ \cline{2-4} 
                                       & F1               & 0.3057 ± 0.0034                & \textbf{0.3243 ± 0.0013}        \\ \hline
\multirow{3}{*}{\textbf{Googlemap CT}} & Precision        & 0.6299 ± 0.0987                & \textbf{0.6305 ± 0.0057}        \\ \cline{2-4} 
                                       & Recall           & 0.6845 ± 0.0003                & \textbf{0.6905 ± 0.0002}        \\ \cline{2-4} 
                                       & F1               & 0.6256 ± 0.0034                & \textbf{0.6354 ± 0.0012}        \\ \hline
\multirow{3}{*}{\textbf{Stack elec}}   & Precision        & 0.6176 ± 0.1345                & \textbf{0.6565 ± 0.0056}        \\ \cline{2-4} 
                                       & Recall           & 0.6956 ± 0.1575                & \textbf{0.7556 ± 0.0056}        \\ \cline{2-4} 
                                       & F1               & 0.6345 ± 0.1356                & \textbf{0.6796 ± 0.0024}        \\ \hline
\multirow{3}{*}{\textbf{Stack ubuntu}} & Precision        & 0.6257 ± 0.0045                & \textbf{0.6824 ± 0.0054}        \\ \cline{2-4} 
                                       & Recall           & 0.7133 ± 0.0023                & \textbf{0.7554 ± 0.1224}        \\ \cline{2-4} 
                                       & F1               & 0.6894 ± 0.0876                & \textbf{0.7156 ± 0.0056}        \\ \hline
\end{tabular}
\caption{Ablation experiment on the temporal encoding for the edge classification performance of \projshort}
\end{table*}

\begin{table*}[!t]
\centering
\begin{tabular}{c|c|c|c|c}
\hline
\textbf{Metric}          & \textbf{Task Type}   & \textbf{Datasets} & \textbf{w/o Temporal Encoding} & \textbf{with Temporal Encoding} \\ \hline
\multirow{8}{*}{ROC-AUC} & \multirow{4}{*}{tr.} & Enron             & 0.9112 ± 0.0563                & \textbf{0.9887 ± 0.001}         \\ \cline{3-5} 
                         &                      & ICEWS1819         & 0.9224 ± 0.4568                & \textbf{0.9950 ± 0.0014}        \\ \cline{3-5} 
                         &                      & Googlemap CT      & 0.8643 ± 0.0654                & \textbf{0.9127 ± 0.0037}        \\ \cline{3-5} 
                         &                      & GDELT             & 0.9123 ± 0.1455                & \textbf{0.9998 ± 0.0001}        \\ \cline{2-5} 
                         & \multirow{4}{*}{in.} & Enron             & 0.8765 ± 0.0674                & \textbf{0.9367 ± 0.0011}        \\ \cline{3-5} 
                         &                      & ICEWS1819         & 0.8854 ± 0.0545                & \textbf{0.9543 ±. 0.0013}       \\ \cline{3-5} 
                         &                      & Googlemap CT      & 0.8223 ± 0.1676                & \textbf{0.8857 ± 0.2327}        \\ \cline{3-5} 
                         &                      & GDELT             & 0.8675 ± 0.0664                & \textbf{0.9577 ± 0.0012}        \\ \hline
\multirow{8}{*}{AP}      & \multirow{4}{*}{tr.} & Enron             & 0.9254 ± 0.0854                & \textbf{0.9943 ± 0.0001}        \\ \cline{3-5} 
                         &                      &  ICEWS1819                  & 0.9346 ± 0.0056                & \textbf{0.9970 ± 0.0012}        \\ \cline{3-5} 
                         &                      & Googlemap CT      & 0.8534 ± 0.0885                & \textbf{0.9144 ± 0.0013}        \\ \cline{3-5} 
                         &                      & GDELT             & 0.9245 ± 0.0878                & \textbf{0.9776 ± 0.0034}        \\ \cline{2-5} 
                         & \multirow{4}{*}{in.} & Enron             & 0.8956 ± 0.0945                & \textbf{0.9550 ± 0.0013}        \\ \cline{3-5} 
                         &                      & ICEWS1819         & 0.8984 ± 0.0044                & \textbf{0.9634 ± 0.0010}        \\ \cline{3-5} 
                         &                      & Googlemap CT      & 0.8445 ± 0.0845                & \textbf{0.8890 ± 0.1134}        \\ \cline{3-5} 
                         &                      & GDELT             & 0.8456 ± 0.0876                & \textbf{0.9369 ± 0.0056}        \\ \hline
\end{tabular}
\caption{Ablation experiment on the temporal encoding for the future link prediction performance of \proj}
\end{table*}

\textbf{Impact of Temporal Encoding on Future Link Prediction and Edge Classification:} To understand the impact of temporal encoding, we run experiments including and excluding the temporal encoding as the edge attribute into the input of student-model. We observe that the performance of \proj on both future link prediction and the edge classification task drops by the exclusion of temporal encoding. However, the performance degradation on the future link prediction task is more significant than the edge classification task. Without the presence of time information affects the proper prediction of future edges more compared to the edge classification where the semantic information from the knowledge distillation term might help \proj to perform comparatively better.






\end{document}